\documentclass{article}

\usepackage{amssymb,amsmath}

\usepackage{microtype}
\usepackage{graphicx}
\usepackage{subfigure}
\usepackage{booktabs} 

\usepackage{hyperref}

\usepackage[accepted]{icml2020}

\icmltitlerunning{When Does Self-Supervision Help Graph Convolutional Networks?}

\usepackage[small,bf]{caption}
\usepackage{multirow}
\definecolor{gray}{rgb}{0.5, 0.5, 0.5}

\usepackage{makecell}

\usepackage{arydshln}

\begin{document}

\twocolumn[
\icmltitle{When Does Self-Supervision Help Graph Convolutional Networks?
}



\icmlsetsymbol{equal}{*}

\begin{icmlauthorlist}
\icmlauthor{Yuning You}{equal,tamu}
\icmlauthor{Tianlong Chen}{equal,tamu}
\icmlauthor{Zhangyang Wang}{tamu}
\icmlauthor{Yang Shen}{tamu}
\end{icmlauthorlist}

\icmlaffiliation{tamu}{Texas A\&M University}

\icmlcorrespondingauthor{Yang Shen}{yshen@tamu.edu}

\icmlkeywords{Machine Learning, ICML}

\vskip 0.3in
]



\printAffiliationsAndNotice{\icmlEqualContribution} 

\newcommand{\red}[1]{\textcolor{red}{#1}}
\newcommand{\blue}[1]{\textcolor{blue}{#1}}
\newcommand{\purple}[1]{\textcolor{purple}{#1}}
\newcommand{\ts}{\textsuperscript}
\definecolor{darkred}{RGB}{192, 0, 0}
\newcommand{\darkred}[1]{\textcolor{darkred}{#1}}

\begin{abstract}
    \vspace{-0.3em}
Self-supervision as an emerging technique has been employed to train convolutional neural networks (CNNs) for more transferrable, generalizable, and robust representation learning of images. Its introduction to graph convolutional networks (GCNs) operating on graph data is however rarely explored.  In this study, we report the first systematic exploration and assessment of incorporating self-supervision into  GCNs.  We first elaborate three mechanisms to incorporate
self-supervision into GCNs, analyze the limitations of pretraining \& finetuning and self-training, and proceed to focus on multi-task learning.
Moreover, we propose to investigate three
novel self-supervised learning tasks for GCNs with theoretical rationales and numerical comparisons.  
Lastly, we further integrate multi-task self-supervision into graph adversarial training.
Our results show that, with properly designed task forms and incorporation mechanisms, self-supervision benefits GCNs in gaining more generalizability
and robustness.
Our codes are available at \url{https://github.com/Shen-Lab/SS-GCNs}.
    \vspace{-0.5em}
\end{abstract}
    \vspace{-1em}
\section{Introduction}
    \vspace{-0.3em}
\label{introduction}
Graph convolutional networks (GCNs) \cite{kipf2016semi} generalize convolutional neural networks (CNNs) \cite{lecun1995convolutional} to graph-structured data and exploit the properties of graphs. They have outperformed traditional approaches in numerous graph-based tasks such as node or link classification \cite{kipf2016semi,velivckovic2017graph,qu2019gmnn,verma2019graphmix,karimi2019explainable,you2020l2}, link prediction \cite{zhang2018link}, and graph classification \cite{ying2018hierarchical,xu2018powerful}, 
many of which are   \textit{semi-supervised learning} tasks.
In this paper, we mainly focus our discussion on transductive semi-supervised node classification, as a representative testbed for GCNs, where there are abundant unlabeled nodes and a small number of labeled nodes in the graph,
with the target to predict the labels of remaining unlabeled nodes. 


In a parallel note, self-supervision has raised a surge of interest in the computer vision domain \cite{goyal2019scaling,kolesnikov2019revisiting,mohseni2020self} to make use of rich unlabeled data.
It aims to assist the model to learn more transferable and generalized representation from unlabeled data via pretext tasks, through pretraining (followed by finetuning), or multi-task learning.
The pretext tasks shall be carefully designed in order to facilitate the network to learn downstream-related semantics features \cite{su2019does}.
A number of pretext tasks have been proposed for CNNs, including rotation \cite{gidaris2018unsupervised}, exemplar \cite{dosovitskiy2014discriminative}, jigsaw \cite{noroozi2016unsupervised} and relative patch location prediction  \cite{doersch2015unsupervised}.
Lately, \citet{hendrycks2019using} demonstrated the promise of self-supervised learning as auxiliary regularizations for improving robustness and uncertainty estimation. \citet{chen2020adversarial} introduced adversarial training into self-supervision, to provide the first general-purpose robust pretraining.

In short, GCN tasks usually admit transductive semi-supervised settings, with tremendous unlabeled nodes; meanwhile, self-supervision plays an increasing role in utilizing unlabeled data in CNNs. In view of the two facts, we are naturally motivated to ask the following interesting, yet rarely explored question:
\begin{itemize}
\vspace{-0.5em}
    \item[] \textit{Can self-supervised learning play a similar role in GCNs to improve their generalizability and robustness}?\vspace{-0.5em}
\end{itemize}
\textbf{Contributions.} This paper presents the first systematic study on how to incorporate self-supervision in GCNs, unfolded by addressing three concrete questions:
\begin{itemize}
    \vspace{-0.3em}
	\item[Q1:]
	Could GCNs benefit from self-supervised learning in their classification performance?
	If yes, how to incorporate it in GCNs to maximize the gain?
	    \vspace{-0.2em}
	\item[Q2:]
	Does the design of pretext tasks matter? What are the useful self-supervised pretext tasks for GCNs?
	    \vspace{-0.2em}
	\item[Q3:]
	Would self-supervision also affect the adversarial robustness of GCNs? If yes, how to design pretext tasks?\vspace{-0.3em}
\end{itemize}
Directly addressing the above questions, our contributions are summarized as follows:
\begin{itemize}
    \vspace{-0.3em}
	\item[A1:]
	We demonstrate the effectiveness of incorporating self-supervised learning in GCNs through \textit{multi-task learning}, i.e. as a regularization term in GCN training.
	It is compared favorably against self-supervision as pretraining, or via self-training \cite{sun2019multi}.
    \vspace{-0.2em}
	\item[A2:]
   We investigate \textit{three self-supervised tasks} based on graph properties. Besides the node clustering task previously mentioned in \cite{sun2019multi}, we propose two new types of tasks: graph partitioning and completion.  We further illustrate that different models and datasets seem to prefer different self-supervised tasks.\vspace{-0.2em}
	\item[A3:]
	We further generalize the above findings into the \textit{adversarial training} setting. We provide extensive results to show that self-supervision also improves robustness of GCN under various attacks, without requiring larger models nor additional data.
\end{itemize}

\section{Related Work}
\label{related_work}
\textbf{Graph-based semi-supervised learning.}
Semi-supervised graph-based learning works with the crucial assumption that the nodes connected with edges of larger weights are more likely to have the same label \cite{zhu2009introduction}.
There are abundance of work on graph-based methods, e.g. (randomized) mincuts \cite{blum2001learning,blum2004semi}, Boltzmann machines \cite{getz2006semi,zhu2002towards} and graph random walks \cite{azran2007rendezvous,szummer2002partially}.
Lately, graph convolutional network (GCN) \cite{kipf2016semi} and its variants \cite{velivckovic2017graph,qu2019gmnn,verma2019graphmix} have gained their popularity by extending the assumption from a hand-crafted one to a data-driven fashion. A detailed review could be referred to \cite{wu2019comprehensive}.

\textbf{Self-supervised learning.}
Self-supervision is a promising direction for neural networks to learn more transferable, generalized and robust features in computer vision domain \cite{goyal2019scaling,kolesnikov2019revisiting,hendrycks2019using}. So far, the usage of self-supervision in CNNs mainly falls under two categories: \textit{pretraining \& finetuning}, or \textit{multi-task learning}. In pretraining \& finetuning. the CNN is first pretrained with self-supervised pretext tasks, and then finetuned with the target task supervised by labels \cite{trinh2019selfie,noroozi2016unsupervised,gidaris2018unsupervised},
while in multi-task learning the network is trained simultaneously with a joint objective of the target supervised task and the self-supervised task(s). \cite{doersch2017multi,ren2018cross}.

To our best knowledge, there has been only one recent work pursuing self-supervision in GCNs \cite{sun2019multi}, where a node clustering task is adopted through self-training.  However, self-training suffers from limitations including performance ``saturation" and degrading (to be detailed in Sections \ref{ss_schemes} and \ref{ss_generalizability} for theoretical rationales and empirical results).  It also restricts the types of self-supervision tasks that can be incorporated.  




\textbf{Adversarial attack and defense on graphs.}
Similarly to CNNs, 
the wide applicability and vulnerability of GCNs raise an urgent demand for improving their robustness.
Several algorithms are proposed to attack and defense on graph \cite{dai2018adversarial,zugner2018adversarial,wang2019adversarial,wu2019adversarial,wang2019graphdefense}.

\citet{dai2018adversarial} developed attacking methods by dropping edges, based on gradient descent, genetic algorithms and reinforcement learning.
\citet{zugner2018adversarial} proposed an FSGM-based approach to attack the edges and features.
Lately, more diverse defense approaches emerge.
\citet{dai2018adversarial} defended the adversarial attacks by directly training on perturbed graphs.
\citet{wu2019adversarial} gained robustness by learning graphs from the continuous function.
\citet{wang2019adversarial} used graph refining and adversarial contrasting learning to boost the model robustness.
\citet{wang2019graphdefense} proposed to involve unlabeled data with pseudo labels that enhances scalability to large graphs.

\section{Method}
\label{method}
In this section, we first elaborate three candidate schemes to incorporate  self-supervision with GCNs. We then design novel self-supervised tasks, each with its own rationale explained. Lastly we generalize self-supervised to GCN adversarial defense.

\subsection{Graph Convolutional Networks}
Given an undirected graph $\mathcal{G} = \{ \mathcal{V}, \mathcal{E} \}$,
where $\mathcal{V} = \{v_1, ..., v_{|\mathcal{V}|}\}$ represents the node set with $|\mathcal{V}|$ nodes,
$\mathcal{E} = \{e_1, ..., e_{|\mathcal{E}|}\}$ stands for the edge set with $|\mathcal{E}|$ edges,
and $e_n = (v_i, v_j)$ indicates an edge between nodes $v_i$ and $v_j$.
Denoting $\boldsymbol{X} \in \mathbb{R}^{|\mathcal{V}| \times N}$ as the feature matrix where $\boldsymbol{x}_n = \boldsymbol{X}[n, :]^T$ is the $N$-dimensional attribute vector of the node $v_n$,
and $\boldsymbol{A} \in \mathbb{R}^{|\mathcal{V}| \times |\mathcal{V}|}$ as the adjacency matrix
where $a_{ij} = \boldsymbol{A}[i,j] = \{ \begin{smallmatrix} 1, \, \text{if} \, (v_i, v_j) \in \mathcal{E} \\ 0, \, \text{otherwise} \end{smallmatrix}$ and $a_{ij} = a_{ji}$, 
the GCN model of semi-supervised classification with two layers \cite{kipf2016semi} is formulated as:
\begin{equation} \label{gcn}
    \boldsymbol{Z} = \hat{\boldsymbol{A}}\,\text{ReLU} (\hat{\boldsymbol{A}} \boldsymbol{X} \boldsymbol{W}_0)\, \boldsymbol{W}_1,
\end{equation}
where $\hat{\boldsymbol{A}} = \tilde{\boldsymbol{D}}^{-\frac{1}{2}} (\boldsymbol{A} + \boldsymbol{I}) \tilde{\boldsymbol{D}}^{-\frac{1}{2}}$, and $\tilde{\boldsymbol{D}}$ is the degree matrix of $\boldsymbol{A} + \boldsymbol{I}$.
Here we do not apply softmax function to the output but treat it as a part of the loss described below.

We can treat  $\hat{\boldsymbol{A}}\, \text{ReLU} (\hat{\boldsymbol{A}} \boldsymbol{X} \boldsymbol{W}_0)$ in \eqref{gcn} as the feature extractor $f_\theta(\boldsymbol{X}, \hat{\boldsymbol{A}})$ of GCNs in general. The parameter set $\theta=\{\boldsymbol{W}_0\}$ in \eqref{gcn} but could include additional parameters for corresponding network architectures in GCN variants \cite{velivckovic2017graph,qu2019gmnn,verma2019graphmix}.  Thus GCN is decomposed into feature extraction and linear transformation as $\boldsymbol{Z} = f_\theta(\boldsymbol{X}, \hat{\boldsymbol{A}}) \boldsymbol{\Theta}$ where parameters $\theta$ and $\boldsymbol{\Theta} = \boldsymbol{W}_1$ are learned from data.  Considering the transductive semi-supervised task, we are provided the labeled node set $\mathcal{V}_\mathrm{label} \subset \mathcal{V}$ with $|\mathcal{V}_\mathrm{label}| \ll |\mathcal{V}|$
and the label matrix $\boldsymbol{Y} \in \mathbb{R}^{|\mathcal{V}| \times N^\prime}$ with label dimension $N^\prime$ (for a classification task $N^\prime = 1$). 
Therefore, the model parameters in GCNs are learned by minimizing the supervised loss calculated between the output and the true label for labeled nodes, which can be formulated as:
\begin{align}
    \boldsymbol{Z} & = f_\theta(\boldsymbol{X}, \hat{\boldsymbol{A}}) \boldsymbol{\Theta}, \notag \\
    \theta^*, \boldsymbol{\Theta}^* & = \arg\min_{\theta, \boldsymbol{\Theta}} \mathcal{L}_\mathrm{sup}(\theta, \boldsymbol{\Theta})  \notag \\
    & = \arg\min_{\theta, \boldsymbol{\Theta}} \frac{1}{|\mathcal{V}_\mathrm{label}|} \sum_{v_n \in \mathcal{V}_\mathrm{label}} L(\boldsymbol{z}_n, \boldsymbol{y}_n), \label{gcn_variants}
\end{align}
where $L(\cdot,\cdot)$ is the loss function for each example, $\boldsymbol{y}_n = \boldsymbol{Y}[n, :]^T$ is the annotated label vector, and $\boldsymbol{z}_n = \boldsymbol{Z}[n, :]^T$ is the true label vector for $v_n \in \mathcal{V}_\mathrm{label}$.  

\begin{figure*}[t]
\begin{center}
  \includegraphics[width=0.8\linewidth]{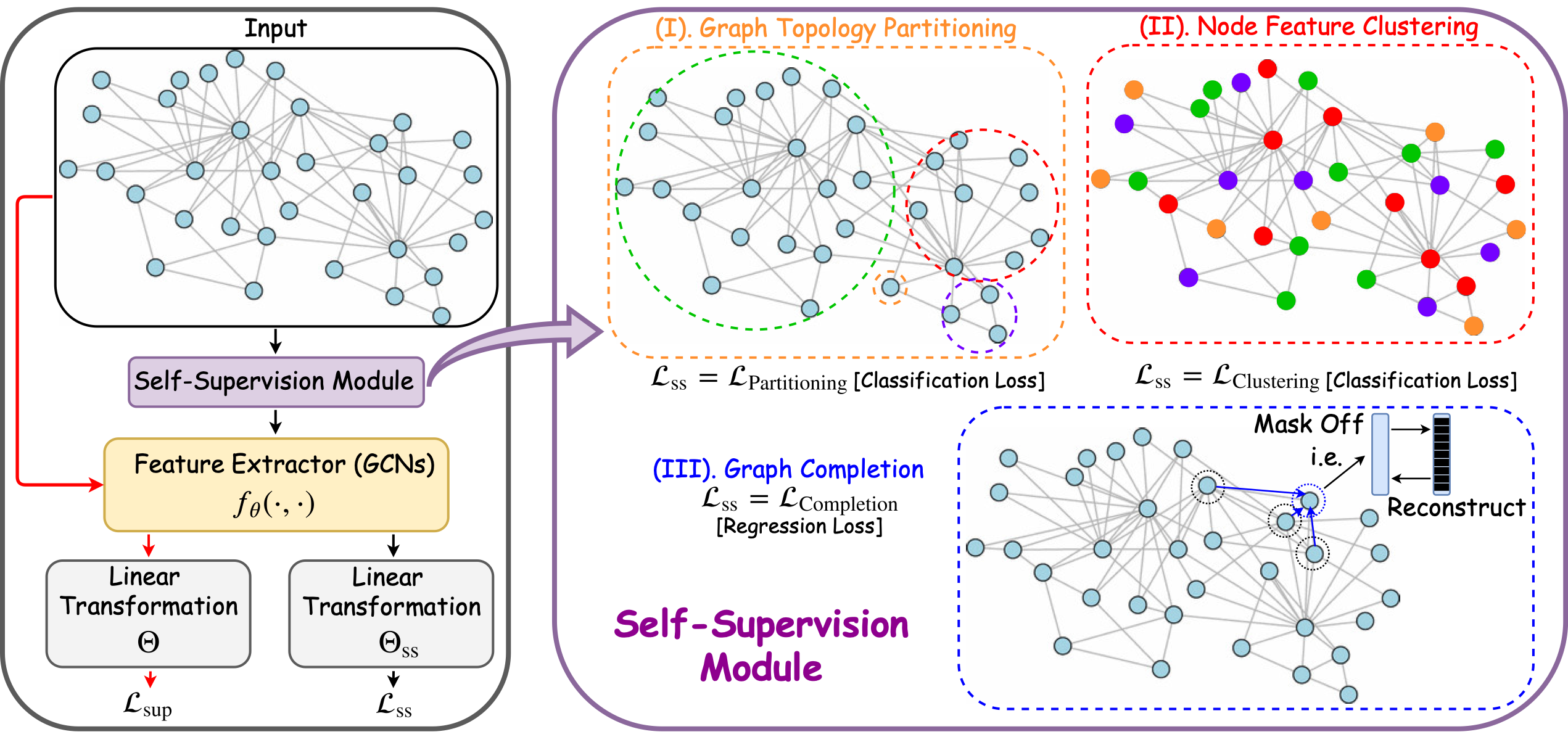}
\end{center}
  \caption{The overall framework for self-supervision on GCN through \textit{multi-task learning}. The target task and auxiliary self-supervised tasks share the same feature extractor $f_\theta(\cdot, \cdot)$ with their individual linear transformation parameters $\boldsymbol{\Theta}, \boldsymbol{\Theta}_\mathrm{ss}$.}  
\label{fig:framework}
\end{figure*}
\subsection{Three Schemes: Self-Supervision Meets GCNs} \label{ss_schemes}
Inspired by relevant discussions in CNNs \cite{goyal2019scaling,kolesnikov2019revisiting,hendrycks2019using}, we next investigate three possible schemes to equip a GCN
with a self-supervised task (``ss''), given the input $\boldsymbol{X}_\mathrm{ss}, \hat{\boldsymbol{A}}_\mathrm{ss}$
the label $\boldsymbol{Y}_\mathrm{ss}$ and the node set $\mathcal{V}_\mathrm{ss}$.

\textbf{Pretraining \& finetuning.}
In the pretraining process, the network is trained with the self-supervised task as following:
\begin{align}
    \boldsymbol{Z}_\mathrm{ss} & = f_\theta(\boldsymbol{X}_\mathrm{ss}, \hat{\boldsymbol{A}}_\mathrm{ss}) \boldsymbol{\Theta}_\mathrm{ss}, \notag \\
    \theta^*_\mathrm{ss}, \boldsymbol{\Theta}^*_\mathrm{ss} & = \arg\min_{\theta, \boldsymbol{\Theta}_\mathrm{ss}} \mathcal{L}_\mathrm{ss}(\theta, \boldsymbol{\Theta}_\mathrm{ss})  \notag \\
    & = \arg\min_{\theta, \boldsymbol{\Theta}} \frac{1}{|\mathcal{V}_\mathrm{ss}|} \sum_{v_n \in \mathcal{V}_\mathrm{ss}} L_\mathrm{ss}(\boldsymbol{z}_{\mathrm{ss},n}, \boldsymbol{y}_{\mathrm{ss},n}), \label{gcn_variants_ss_pretraining}
\end{align}
where $\boldsymbol{\Theta}_\mathrm{ss}$ is the linear transformation parameter, and $L_\mathrm{ss}(\cdot, \cdot)$ is the loss function of the self-supervised task, $\boldsymbol{z}_{\mathrm{ss},n} = \boldsymbol{Z}_\mathrm{ss}[n,:]^T, \boldsymbol{y}_{\mathrm{ss},n} = \boldsymbol{Y}_\mathrm{ss}[n,:]^T$.
Then in the finetuing process the feature extractor $f_\theta(\cdot, \cdot)$ is trained in formulation \eqref{gcn_variants} using $\theta^*_\mathrm{ss}$ to initialize parameters $\theta$.
\begin{table}[ht]
\scriptsize
\begin{center}
\caption{Comparing performances of GCN through pretraining \& finetuning (P\&F) and multi-task learning (MTL) with graph partitioning (see Section \ref{ss_tasks}) on the PubMed dataset. Reported numbers correspond to classification accuracy in percent.}
\label{tab_experiment_pf_mt}
\begin{tabular}{c | c | c | c}
    \hline
    \hline
    Pipeline & GCN & P\&F & MTL \\
    \hline
    \hline
    Accuracy & 79.10 $\pm$ 0.21 & 79.19 $\pm$ 0.21 & 80.00 $\pm$ 0.74 \\
    \hline
\end{tabular}
\end{center}
\end{table}

Pretraining \& finetuning is arguably the most straightforward option for self-supervision benefiting GCNs.
However, our preliminary experiment found little performance gain from it on a large dataset Pubmed (Table \ref{tab_experiment_pf_mt}).  We conjecture that it is due to (1) ``switching'' to a different objective function $\mathcal{L}_\mathrm{sup}(\cdot,\cdot)$ in finetuning from that in pretraining $\mathcal{L}_\mathrm{ss}(\cdot,\cdot)$; and (2) training a shallow GCN in the transductive semi-supervised setting, which was shown to beat deeper GCNs causing over-smoothing or ``information loss'' \cite{li2018deeper,oonograph}.  We will systematically assess and analyze this scheme over multiple datasets and combined with other self-supervision tasks in Section \ref{ss_generalizability}.  



\textbf{Self-training.}
\cite{sun2019multi} is the only prior work that pursues self-supervision in GCNs and it does so through self-training. With both labeled and unlabeled data, a typical \textit{self-training} pipeline starts by pretraining a model over the labeled data, then assigning ``pseudo-labels'' to highly confident unlabeled samples, and including them into the labeled data for the next round of training.  The process could be repeated several rounds and can be formulated in each round similar to formulation \eqref{gcn_variants} with $\mathcal{V}_\mathrm{label}$ updated. The authors of \cite{sun2019multi} proposed a multi-stage self-supervised (M3S) training algorithm, where self-supervision was injected to align and refine the pseudo labels for the unlabeled nodes.
\begin{table}[ht]
\scriptsize
\begin{center}
\caption{Experiments for GCN through M3S. \textcolor{gray}{Gray} numbers are from \cite{sun2019multi}.}
\label{tab_experiment_self_training}
\begin{tabular}{c | c | c | c}
    \hline
    \hline
    Label Rate & 0.03\% & 0.1\% & 0.3\% (Conventional dataset split) \\
    \hline
    \hline
    GCN & \textcolor{gray}{51.1} & \textcolor{gray}{67.5} & 79.10 $\pm$ 0.21 \\
    M3S & \textcolor{gray}{59.2} & \textcolor{gray}{70.6} & 79.28 $\pm$ 0.30 \\
    \hline
\end{tabular}
\end{center}
\end{table}

Despite improving performance in previous few-shot experiments, M3S shows  performance gain ``saturation'' in Table \ref{tab_experiment_self_training} as the label rate grows higher, 
echoing literature \cite{zhu2009introduction,li2018deeper}. Further, we will show and rationalize their limited performance boost in Section \ref{ss_generalizability}.

\textbf{Multi-task learning.}
Considering a target task and a self-supervised task for a GCN with \eqref{gcn_variants}, the output and the training process can be formulated as:
\begin{align}
    \boldsymbol{Z} & = f_\theta(\boldsymbol{X}, \hat{\boldsymbol{A}}) \boldsymbol{\Theta}, \quad
    \boldsymbol{Z}_\mathrm{ss} = f_\theta(\boldsymbol{X}_\mathrm{ss}, \hat{\boldsymbol{A}}_\mathrm{ss}) \boldsymbol{\Theta}_\mathrm{ss}, \notag  \\
    \theta^*, \boldsymbol{\Theta}^*, \boldsymbol{\Theta}^*_\mathrm{ss} & = \arg\min_{\theta, \boldsymbol{\Theta},\boldsymbol{\Theta}_\mathrm{ss}} \alpha_1 \mathcal{L}_\mathrm{sup}(\theta, \boldsymbol{\Theta}) + \alpha_2 \mathcal{L}_\mathrm{ss}(\theta, \boldsymbol{\Theta}_\mathrm{ss}),  
    \label{ss_gcn_variants}
\end{align}
where $\alpha_1, \alpha_2 \in \mathbb{R}_{>0}$ are the weights for the overall supervised loss $\mathcal{L}_\mathrm{sup}(\theta, \boldsymbol{\Theta})$ as defined in \eqref{gcn_variants} and those for the self-supervised loss  $\mathcal{L}_\mathrm{ss}(\theta, \boldsymbol{\Theta}_\mathrm{ss})$ as defined in \eqref{gcn_variants_ss_pretraining}, respectively. To optimize the weighted sum of their losses, the target supervised and self-supervised tasks share the same feature extractor $f_\theta(\cdot,\cdot)$ but have their individual linear transformation parameters $\boldsymbol{\Theta}^*$ and $\boldsymbol{\Theta}^*_\mathrm{ss}$ as in Figure \ref{fig:framework}. 

In the problem \eqref{ss_gcn_variants}, we regard the self-supervised task as a \textit{regularization} term throughout the network training.
The regularization term is traditionally and widely used in graph signal processing, and a famous one is graph Laplacian regularizer (GLR) \cite{shuman2013emerging,bertrand2013seeing,milanfar2012tour,sandryhaila2014big,wu2016estimating} which penalizes incoherent (i.e. nonsmooth) signals across adjacent nodes \cite{chen2017bias}.
Although the effectiveness of GLR has been shown in graph signal processing, the regularizer is manually set simply following the smoothness prior without the involvement of data, whereas the self-supervised task acts as the regularizer learned from unlabeled data under the minor guidance of human prior.
Therefore, a properly designed task would introduce data-driven prior knowledge that improves the model generalizability, as show in Table \ref{tab_experiment_pf_mt}.

In total, multi-task learning is the most general framework among the three. Acting as the data-driven regularizer during training, it makes no assumption on the self-supervised task type.  It is also experimentally verified to be the most effective among all the three (Section \ref{experimentss}).

\subsection{GCN-Specific Self-Supervised Tasks} \label{ss_tasks}

While Section \ref{ss_schemes} discusses the ``mechanisms" by which GCNs could be trained with self-supervision, here we expand a ``toolkit'' of self-supervised tasks for GCNs. 
We show that, by utilizing the rich node and edge information in a graph, a variety of GCN-specific self-supervised tasks (as summarized in Table \ref{tab_overview_ss_tasks}) could be defined and will be further shown to benefit various types of supervised/downstream tasks. They will assign different pseudo-labels to unlabeled nodes and solve formulation in \eqref{ss_gcn_variants}. 
\begin{table}[ht]
\scriptsize
\begin{center}
\caption{Overview of three self-supervised tasks.}
\label{tab_overview_ss_tasks}
\resizebox{0.48\textwidth}{!}{
\begin{tabular}{c | c | c | c}
    \hline
    \hline
    Task & Relied Feature & Primary Assumption & Type \\
    \hline
    \hline
    Clustering & Nodes & Feature Similarity & Classification \\
    Partitioning & Edges & Connection Density & Classification \\
    Completion & Nodes \& Edges & Context based Representation & Regression \\
    \hline
\end{tabular}}
\end{center}
\end{table}


\textbf{Node clustering.}
Following M3S \cite{sun2019multi},
one intuitive way to construct a self-supervised task is via the node clustering algorithm.
Given the node set $\mathcal{V}$ with the feature matrix $\boldsymbol{X}$ as input, with a preset number of clusters  $K \in \{1, \ldots, |\mathcal{V}|\}$ (treated as a hyperparameter in our experiments), the clustering algorithm will output a set of node sets $\{\mathcal{V}_{\mathrm{clu},1}, ..., \mathcal{V}_{\mathrm{clu},K} | \mathcal{V}_{\mathrm{clu},n} \subseteq \mathcal{V}, \ n=1,...,K\}$ 
such that:
\begin{gather*}
    \mathcal{V}_{\mathrm{clu},n} \neq \emptyset \quad (n = 1,\ldots,K), \quad 
    \cup^K_{n=1} \mathcal{V}_{\mathrm{clu},n} = \mathcal{V}, \\
    \mathcal{V}_{\mathrm{clu},i} \cap \mathcal{V}_{\mathrm{clu},j} = \emptyset \quad (\forall i,j = 1,...,K \thickspace \text{and} \thickspace i \neq j).
\end{gather*}
With the clusters of node sets, we assign cluster indices as self-supervised labels to all the nodes:
\begin{equation} \label{clustering}
    y_{\mathrm{ss},n} = k \thickspace \text{if} \thickspace v_n \in \mathcal{V}_{\mathrm{clu},k}\ \   (\forall n = 1,\ldots,|\mathcal{V}|, \,\forall k = 1,\ldots,K). \notag
\end{equation}

\textbf{Graph partitioning.}
Clustering-related algorithms are node feature-based, with the rationale of grouping nodes with similar attributes.
Another rationale to group nodes can be based on topology in graph data.  In particular two nodes connected by a ``strong'' edge (with a large weight) are highly likely of the same label class \cite{zhu2009introduction}.
Therefore, we propose a topology-based self-supervision using graph partitioning.  

Graph partitioning is to partition the nodes of a graph into roughly equal subsets, such that the number of edges connecting nodes across subsets is minimized \cite{karypis1995multilevel}.
Given the node set $\mathcal{V}$, the edge set $\mathcal{E}$ and the adjacency matrix $\boldsymbol{A}$ as the input, with a preset number of partitions $K \in \{1, \ldots, |\mathcal{V}|\}$ (a hyperparameter in our experiments), a graph partitioning algorithm will output a set of node sets $\{\mathcal{V}_{\mathrm{par},1}, \ldots, \mathcal{V}_{\mathrm{par},K} | \mathcal{V}_{\mathrm{par},n} \subseteq \mathcal{V}, \ n=1,\ldots,K\}$ such that:
\begin{gather*}
    \mathcal{V}_{\mathrm{par},n} \neq \emptyset \quad (\forall n = 1,...,K),\quad 
    \cup^K_{n=1} \mathcal{V}_{\mathrm{par},n} = \mathcal{V}, \\
    \mathcal{V}_{\mathrm{par},i} \cap \mathcal{V}_{\mathrm{par},j} = \emptyset \quad (\forall i,j = 1,...,K \thickspace \text{and} \thickspace i \neq j), 
\end{gather*}
which is similar to the case of node clustering.  
In addition, balance constraints are enforced for graph partitioning 
\big($K \frac{\text{max}_k|\mathcal{V}_{\mathrm{par},k}|}{|\mathcal{V}|} \leqslant 1 + \epsilon,  \text{where } \epsilon \in (0, 1)$ \big) and the objective of graph partitioning is to minimize the edgecut \big($\text{edgecut} = \frac{1}{2} \sum_{k=1}^K \sum_{v_i \in \mathcal{V}_{\mathrm{par},k}} \sum_{\substack{(v_i, v_j) \in \mathcal{E}, \\ \text{and} \thickspace v_j \notin \mathcal{V}_{\mathrm{par},k}}} a_{ij}$\big). 

With the node set partitioned along with the rest of the graph, we assign partition indices as self-supervised labels: $
    y_{\mathrm{ss},n} = k \thickspace \text{if} \thickspace v_n \in \mathcal{V}_{\mathrm{par},k}, n = 1,...,|\mathcal{V}|, \forall k = 1,\ldots,K. $


Different from node clustering based on node features, graph partitioning provides the prior regularization based on graph topology, which is similar to graph Laplacian regularizer (GLR) \cite{shuman2013emerging,bertrand2013seeing,milanfar2012tour,sandryhaila2014big,wu2016estimating} that also adopts the idea of ``connection-prompting similarity".
However, GLR, which is already injected into the GCNs architecture, locally smooths all nodes with their neighbor nodes. In contrast, graph partitioning considers global smoothness by utilizing all connections to group nodes with heavier connection densities.

\begin{figure}[t]
\begin{center}
  \includegraphics[width=1\linewidth]{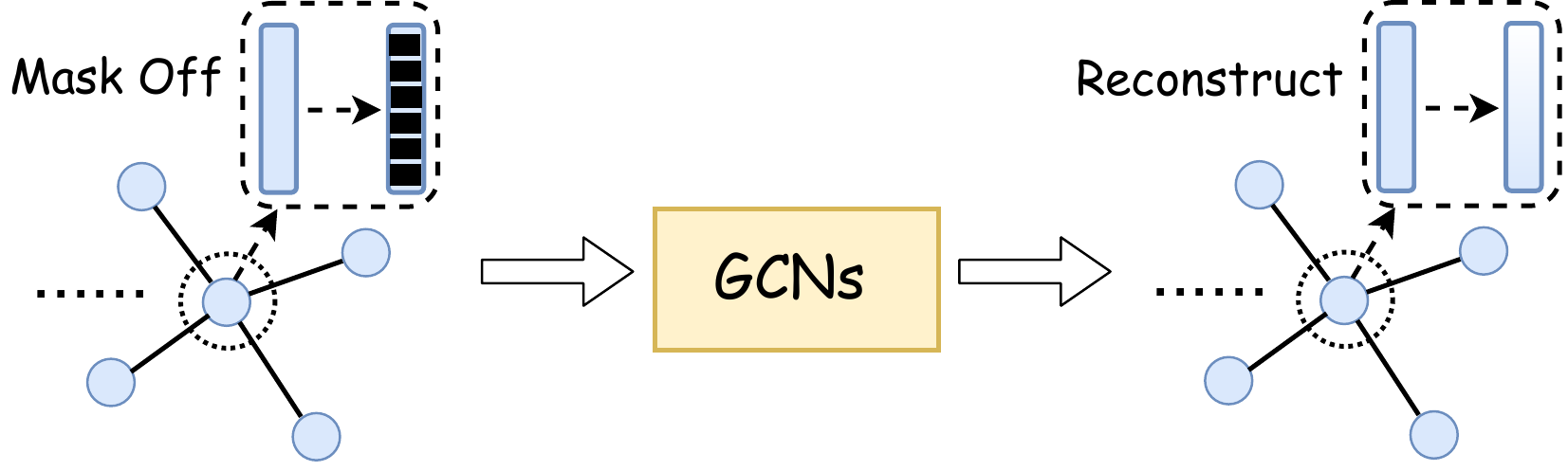}
\end{center}
  \caption{Graph completion for a target node. With the target-node feature masked and neighbors' features and connections provided, GCNs will recover the masking feature based on the neighborhood information.} 
\label{fig_inpainting}
\end{figure}

\textbf{Graph completion.}
Motivated by image inpainting a.k.a. completion \cite{yu2018generative} in computer vision (which aims to fill missing pixels of an image), we propose graph completion, a novel regression task, as a  self-supervised task.  
As an analogy to image completion and illustrated in Figure \ref{fig_inpainting}, our graph completion first masks target nodes
by removing their features.  It then aims at recovering/predicting masked node features by feeding to GCNs unmasked node features (currently restricted to second-order neighbors of each target node for 2-layer GCNs).


We design such a self-supervised task for the following reasons:
1) the completion labels are free to obtain, which is the node feature itself; and
2) we consider graph completion can aid the network for better feature representation, which teaches the network to extract feature from the context.

\subsection{Self-Supervision in Graph Adversarial Defense}
With the three self-supervised tasks introduced for GCNs to gain generalizability toward better-performing supervised learning (for instance, node classification), we proceed to examine their possible roles in gaining robustness against various graph adversarial attacks.  

\textbf{Adversarial attacks.}
We focus on single-node direct evasion attacks: 
a node-specific attack type on the 
attributes/links of the target node $v_n$ under certain constraints following \cite{zugner2018adversarial}, whereas the trained model (i.e. the model parameters $(\theta^*, \boldsymbol{\Theta}^*)$) remains unchanged during/after the attack.
The attacker $g$ generates perturbed feature and adjacency matrices,  $\boldsymbol{X}^\prime$ and $\boldsymbol{A}^\prime$, as:
\begin{equation}
    \boldsymbol{X}^\prime, \boldsymbol{A}^\prime = g(\boldsymbol{X}, \boldsymbol{A}, \boldsymbol{Y}, v_n, \theta^*, \boldsymbol{\Theta}^*),
\end{equation}
with (attribute, links and label of) the target node and the model parameters as inputs.
The attack can be on links, 
(node) features or links \& features.

\textbf{Adversarial defense.}
An effective approach for adversarial defense, especially in image domain, is through adversarial training which augments training sets with adversarial examples \cite{goodfellow2014explaining}. 
However, it is difficult to generate adversarial examples in graph domain because of low labeling rates in the transductive semi-supervised setting. 
\citet{wang2019graphdefense} thus proposed to utilize unlabeled nodes in generating adversarial examples. Specifically, they trained a GCN as formulated in \eqref{gcn_variants} to assign pseudo labels $\boldsymbol{Y}_\mathrm{pseudo}$ to unlabeled nodes.  Then they randomly chose two disjoint subsets $\mathcal{V}_\mathrm{clean}$ and $\mathcal{V}_\mathrm{attack}$ from the unlabeled node set  and attacked each target node $v_n \in \mathcal{V}_\mathrm{attack}$ to generate perturbed feature and adjacency matrices  $\boldsymbol{X}^\prime$ and $\boldsymbol{A}^\prime$. 

Adversarial training for graph data can then be formulated as both supervised learning for labeled nodes and recovering pseudo labels for unlabeled nodes (attacked and clean):
\begin{align}
    \boldsymbol{Z} & = f_\theta(\boldsymbol{X}, \hat{\boldsymbol{A}}) \boldsymbol{\Theta},\quad
    \boldsymbol{Z}^\prime = f_\theta(\boldsymbol{X}^\prime, \boldsymbol{A}^\prime) \boldsymbol{\Theta}, \notag \\
    \theta^*, \boldsymbol{\Theta}^*  & = \arg\min_{\theta, \boldsymbol{\Theta}}   \bigl( \mathcal{L}_\mathrm{sup}(\theta, \boldsymbol{\Theta}) + \alpha_3 \mathcal{L}_\mathrm{adv}(\theta, \boldsymbol{\Theta}) \bigr),  
    \label{gcn_variants_adv}
\end{align}

where $\alpha_3$ is a weight for the adversarial loss $\mathcal{L}_\mathrm{adv}(\cdot,\cdot)$, $\boldsymbol{y}_{\mathrm{pseudo},n} = \boldsymbol{Y}_\mathrm{pseudo}[n, :]^T$ and $\boldsymbol{z}^\prime_n = \boldsymbol{Z}^\prime[n, :]^T$. 

\textbf{Adversarial defense with self-supervision.}
With self-supervision working in GCNs formulated as in \eqref{ss_gcn_variants} and adversarial training in \eqref{gcn_variants_adv}, we formulate adversarial training with self-supervision as:
\begin{align}
    \boldsymbol{Z} \,\,\, & = f_\theta(\boldsymbol{X}, \hat{\boldsymbol{A}}) \boldsymbol{\Theta}, \quad
    \boldsymbol{Z}^\prime = f_\theta(\boldsymbol{X}^\prime, \boldsymbol{A}^\prime) \boldsymbol{\Theta}, \notag \\
    \boldsymbol{Z}_\mathrm{ss} & = f_\theta(\boldsymbol{X}_\mathrm{ss}, \boldsymbol{A}_\mathrm{ss}) \notag \\
        \theta^*, \boldsymbol{\Theta}^*, \boldsymbol{\Theta}^*_\mathrm{ss}  & = \arg\min_{\theta, \boldsymbol{\Theta},\boldsymbol{\Theta}_\mathrm{ss}}   \bigl( \alpha_1 \mathcal{L}_\mathrm{sup}(\theta, \boldsymbol{\Theta}) \notag \\ 
          & \qquad  + \alpha_2 \mathcal{L}_\mathrm{ss}(\theta, \boldsymbol{\Theta}_\mathrm{ss})  + \alpha_3 \mathcal{L}_\mathrm{adv}(\theta, \boldsymbol{\Theta}) \bigr), 
    \label{gcn_variants_adv_ss}
\end{align}
where the self-supervised loss is introduced into training with the \textit{perturbed} graph data as input
(the self-supervised label matrix $\boldsymbol{Y}_\mathrm{ss}$ is also generated from perturbed inputs).
It is observed in CNNs that self-supervision improves robustness and uncertainty estimation without requiring larger models or additional data \cite{hendrycks2019using}.  
We thus experimentally explore whether that also extends to GCNs.






\section{Experiments} \label{experimentss}
In this section, we extensively assess, analyze, and rationalize the impact of self-supervision on transductive semi-supervised node classification following \cite{kipf2016semi} on the aspects of:
1) the standard performances of GCN \cite{kipf2016semi} with different self-supervision schemes;
2) the standard performances of multi-task self-supervision on three popular GNN architectures --- GCN, graph attention network (GAT) \cite{velivckovic2017graph}, and graph isomorphism network (GIN) \cite{xu2018powerful}; as well as those on two SOTA models for semi-supervised node classification --- graph Markov neural network (GMNN)  \cite{qu2019gmnn} that introduces  statistical relational learning \cite{koller1998probabilistic,friedman1999learning} into its architecture to facilitate training and GraphMix \cite{verma2019graphmix} that uses the Mixup trick; and 
3) the performance of GCN with multi-task self-supervision in adversarial defense.
Implementation details can be found in Appendix A.


\begin{table}[!htb]
\small
\begin{center}
\caption{Dataset statistics. $|\mathcal{V}|$, $|\mathcal{V}_\mathrm{label}|$, $|\mathcal{E}|$, and $N$ denotes the numbers of nodes, numbers of labeled nodes, numbers of edges, and feature dimension per node, respectively.}
\label{tab_dataset}
\begin{tabular}{c c c c c c c c}
    \hline
    \hline
        Dataset & $|\mathcal{V}|$ & $|\mathcal{V}_\mathrm{label}|$ & $|\mathcal{E}|$ & $N$ & Classes \\
    \hline
    \hline
    Cora & 2,780 & 140 & 13,264 & 1,433 & 7 \\
    Citeseer & 3,327 & 120 & 4,732 & 3,703 & 6 \\
    PubMed & 19,717 & 60 & 108,365 & 500 & 3 \\
    \hline
\end{tabular}
\end{center}
\end{table}

\subsection{Self-Supervision Helps Generalizability} \label{ss_generalizability}
\textbf{Self-supervision incorporated into GCNs through various schemes.}
We first examine three schemes (Section \ref{ss_schemes}) to incorporate self-supervision into GCN training: pretraining \& finetuning, self-training (i.e. M3S \cite{sun2019multi}) and multi-task learning.  The hyper-parameters of M3S are set at default values reported in \cite{sun2019multi}.  The differential effects of the three schemes combined with various self-supervised tasks 
are summarized for three datasets in Table \ref{tab_experiments_schemes}, using the target performances (accuracy in node classification).
Each combination of self-supervised scheme and task is run 50 times for each dataset with different random seeds so that the mean and the standard deviation of its performance can be reported.  
\begin{table}[!htb]
\scriptsize
\begin{center}
\caption{Node classification performances (accuracy; unit: \%) when incorporating three self-supervision tasks (Node \textbf{Clu}stering, Graph \textbf{Par}titioning, and Graph \textbf{Comp}letion) into GCNs through various schemes: pretraining \& finetuning (abbr. P\&T),  self-training M3S \cite{sun2019multi}), and multi-task learning (abbr. MTL).  
\darkred{Red} numbers indicate the best two performances with the mean improvement at least 0.8 (where 0.8 is comparable or less than observed standard deviations). In the case of GCN without self-supervision, \textcolor{gray}{gray} numbers indicate the published results.}
\label{tab_experiments_schemes}
\begin{tabular}{c |c | c | c}
    \hline
    \hline
      & Cora & Citeseer & PubMed \\
    \hline
    \hline
    \multirow{2}{*}{GCN} & 81.00 $\pm$ 0.67 & 70.85 $\pm$ 0.70 & 79.10 $\pm$ 0.21 \\
     & \textcolor{gray}{81.5} & \textcolor{gray}{70.3} & \textcolor{gray}{79.0} \\
    \hline
    P\&F-Clu & \darkred{81.83} $\pm$ 0.53 & 71.06 $\pm$ 0.59 & 79.20 $\pm$ 0.22 \\
    P\&F-Par & 81.42 $\pm$ 0.51 & 70.68 $\pm$ 0.81 & 79.19 $\pm$ 0.21 \\
    P\&F-Comp & 81.25 $\pm$ 0.65 & 71.06 $\pm$ 0.55 & 79.19 $\pm$ 0.39 \\ \hline
    M3S & 81.60 $\pm$ 0.51 & \darkred{71.94} $\pm$ 0.83 & 79.28 $\pm$ 0.30 \\
    \hline
    MTL-Clu & 81.57 $\pm$ 0.59 & 70.73 $\pm$ 0.84 & 78.79 $\pm$ 0.36 \\
    MTL-Par & \darkred{81.83} $\pm$ 0.65 & 71.34 $\pm$ 0.69 & \darkred{80.00} $\pm$ 0.74 \\
    MTL-Comp & 81.03 $\pm$ 0.68 & \darkred{71.66} $\pm$ 0.48 & 79.14 $\pm$ 0.28 \\
    \hline
\end{tabular}
\vspace{-1em}
\end{center}
\end{table}

Results in Table \ref{tab_experiments_schemes} first show that, among the three schemes to incorporate self-supervision into GCNs, pretraining \& finetuning provides some performance improvement for the small dataset Cora but does not do so for the larger datasets Citeseer and PubMed.  This conclusion remains valid regardless of the choice of the specific self-supervised task.  The moderate performance boost echos our previous conjecture: although information about graph structure and features is first learned through self-supervision ($\mathcal{L}_\mathrm{ss}$ as in \eqref{gcn_variants_ss_pretraining}) in the pretraining stage, such information may be largely lost during finetuning  while targeting the target supervised loss alone ($\mathcal{L}_\mathrm{sup}$ as in  \eqref{gcn_variants}). The reason for such information loss being particularly observed in GCNs could be that, the shallow GCNs used in the transductive semi-supervised setting can be more easily ``overwritten'' while switching from one objective function to another in finetuning.  

Through the remaining two schemes, GCNs with self-supervision incorporated could see more significant improvements in the target task (node classification) compared to GCN without self-supervision.  In contrast to pretraining and finetuning that switches the objective function after self-supervision in \eqref{gcn_variants_ss_pretraining} and solves a new optimization problem in \eqref{gcn_variants}, both self-training and multi-task learning incorporate self-supervision into GCNs through one optimization problem and both essentially introduce an additional self-supervision loss to the original formulation in \eqref{gcn_variants}.  

Their difference lies in what pseudo-labels are used and how they are generated for unlabeled nodes.  In the case of self-training, the pseudo-labels are the same as the target-task labels and such ``virtual'' labels are assigned to unlabeled nodes based on their proximity to \textit{labeled} nodes in graph embedding.
In the case of multi-task learning, the pseudo-labels are no longer restricted to the target-task labels and can be assigned to \textit{all} unlabeled nodes by exploiting graph structure and node features without labeled data.  And the target supervision and the self-supervision in multi-task learning are still coupled through common graph embedding.  
So compared to self-training, multi-task learning can be more general (in pseudo-labels) and can exploit more in graph data (through regularization).

\textbf{Multi-task self-supervision on SOTAs.}
\textit{Does multi-task self-supervision help SOTA GCNs?} Now that we have established multi-task learning as an effective mechanism to incorporate self-supervision into GCNs, we set out to explore the added benefits of various self-supervision tasks to SOTAs through multi-task learning. 
Table \ref{tab_experiments} shows that different self-supervised tasks could benefit different network architectures on different datasets to different extents.

\begin{table}[!htb]
\scriptsize
\begin{center}
\caption{Experiments on SOTAs (GCN, GAT, GIN, GMNN, and GraphMix) with multi-task self-supervision.
\darkred{Red} numbers indicate the best two performances for each SOTA.}
\label{tab_experiments}
\begin{tabular}{c | c | c | c}
    \hline
    \hline
     Datasets & Cora & Citeseer & PubMed \\
    \hline
    \hline
    GCN & 81.00 $\pm$ 0.67 & 70.85 $\pm$ 0.70 & 79.10 $\pm$ 0.21 \\
    \hdashline
    GCN+Clu & \darkred{81.57} $\pm$ 0.59 & 70.73 $\pm$ 0.84 & 78.79 $\pm$ 0.36 \\
    GCN+Par & \darkred{81.83} $\pm$ 0.65 & \darkred{71.34} $\pm$ 0.69 & \darkred{80.00} $\pm$ 0.74 \\
    GCN+Comp & 81.03 $\pm$ 0.68 & \darkred{71.66} $\pm$ 0.48 & \darkred{79.14} $\pm$ 0.28 \\
    \hline
    GAT & 77.66 $\pm$ 1.08 & 68.90 $\pm$ 1.07 & \darkred{78.05} $\pm$ 0.46 \\
    \hdashline
    GAT+Clu & 79.40 $\pm$ 0.73 & \darkred{69.88} $\pm$ 1.13 & 77.80 $\pm$ 0.28 \\
    GAT+Par & \darkred{80.11} $\pm$ 0.84 & 69.76 $\pm$ 0.81 & \darkred{80.11} $\pm$ 0.34 \\
    GAT+Comp & \darkred{80.47} $\pm$ 1.22 & \darkred{70.62} $\pm$ 1.26 & 77.10 $\pm$ 0.67 \\
    \hline
    GIN & 77.27 $\pm$ 0.52 & 68.83 $\pm$ 0.40 & 77.38 $\pm$ 0.59 \\
    \hdashline
    GIN+Clu & \darkred{78.43} $\pm$ 0.80 & \darkred{68.86} $\pm$ 0.91 & 76.71 $\pm$ 0.36 \\
    GIN+Par & \darkred{81.83} $\pm$ 0.58 & \darkred{71.50} $\pm$ 0.44 & \darkred{80.28} $\pm$ 1.34 \\
    GIN+Comp & 76.62 $\pm$ 1.17 & 68.71 $\pm$ 1.01 & \darkred{78.70} $\pm$ 0.69 \\
    \hline
    GMNN & 83.28 $\pm$ 0.81 & 72.83 $\pm$ 0.72 & \darkred{81.34} $\pm$ 0.59 \\
    \hdashline
    GMNN+Clu & \darkred{83.49} $\pm$ 0.65 & \darkred{73.13} $\pm$ 0.72 & 79.45 $\pm$ 0.76 \\
    GMNN+Par & \darkred{83.51} $\pm$ 0.50 & \darkred{73.62} $\pm$ 0.65 & 80.92 $\pm$ 0.77 \\
    GMNN+Comp & 83.31 $\pm$ 0.81 & 72.93 $\pm$ 0.79 & \darkred{81.33} $\pm$ 0.59 \\
    \hline
    GraphMix & \darkred{83.91} $\pm$ 0.63 & 74.33 $\pm$ 0.65 & 80.68 $\pm$ 0.57 \\
    \hdashline
    GraphMix+Clu & 83.87 $\pm$ 0.56 & \darkred{75.16} $\pm$ 0.52 & 79.99 $\pm$ 0.82 \\
    GraphMix+Par & \darkred{84.04} $\pm$ 0.57 & \darkred{74.93} $\pm$ 0.43 & \darkred{81.36} $\pm$ 0.33 \\
    GraphMix+Comp & 83.76 $\pm$ 0.64 & 74.43 $\pm$ 0.72 & \darkred{80.82} $\pm$ 0.54 \\
    \hline
\end{tabular}
\end{center}
\end{table}

\textit{When does multi-task self-supervision help SOTAs and why?} 
We note that graph partitioning is generally beneficial to all three SOTAs (network architectures) on all the three datasets, whereas node clustering do not benefit SOTAs on PubMed.  As discussed in Section \ref{ss_schemes} and above,  multi-task learning introduce self-supervision tasks into the optimization problem in \eqref{ss_gcn_variants} as \textit{the data-driven regularization} and these tasks represent various priors (see Section \ref{ss_tasks}).

(1) Feature-based node clustering assumes that feature similarity implies target-label similarity and can group distant nodes with similar features together.  When the dataset is large and the feature dimension is relatively low (such as PubMed), feature-based clustering could be challenged in providing informative pseudo-labels.  

(2) Topology-based graph partitioning assumes that connections in topology implies similarity in labels, which is  safe for the three datasets that are all citation networks. In addition, graph partitioning as a classification task does not impose the assumption overly strong.  Therefore, the prior represented by graph partitioning can be general and effective to benefit GCNs (at least for the types of the target task and datasets considered).  

(3) Topology and feature-based graph completion assumes the feature similarity or smoothness in small neighborhoods of graphs.
Such a context-based feature representation can greatly improve target performance, especially when the neighborhoods are small (such as Citeseer with the smallest average degree among all three datasets).  However, the regression task can be challenged facing denser graphs with larger neighborhoods and more difficult completion tasks (such as the larger and denser PubMed with continuous features to complete). 
That being said, 
the potentially informative prior from graph completion can greatly benefit other tasks, which is validated later (Section \ref{ss_adversarial}). 

\textit{Does GNN architecture affect multi-task self-supervision?} 
For every GNN architecture/model, all three self-supervised tasks improve its performance for some datasets (except for GMNN on PubMed).  The improvements are more significant for GCN, GAT, and GIN.  We conjecture that data-regularization through various priors could benefit 
these three architectures (especially GCN)
with weak priors to begin with.   
In contrast, GMNN sees little improvement with graph completion.
GMNN introduces statistical relational learning (SRL) into the architecture to model the dependency between vertices and their neighbors.   Considering that graph completion aids context-based representation and acts a somewhat similar role as SRL, the self-supervised and the architecture priors can be similar and their combination may not help.    
Similarly GraphMix introduces a data augmentation method Mixup into the architecture to refine feature embedding, which again mitigates the power of graph completion with overlapping aims. 



 
We also report in Appendix B the results in inductive fully-supervised node classification. Self-supervision leads to modest performance improvements in this case, appearing to be more beneficial in semi-supervised or few-shot learning.

\subsection{Self-Supervision Boosts Adversarial Robustness} \label{ss_adversarial}
\textit{What additional benefits could multi-task self-supervision bring to GCNs},  besides improving the generalizability of graph embedding (Section \ref{ss_generalizability})?  
We additionally perform adversarial experiments on GCN with multi-task self-supervision against Nettack \cite{zugner2018adversarial}, to examine its potential benefit on robustness.

We first generate attacks with the same perturbation intensity ($n_\mathrm{perturb} = 2$, see details in Appendix A) as in adversarial training to see the robust generalization. For each self-supervised task, the hyper-parameters are set at the same values as in Table \ref{tab_experiments}. 
Each experiment is repeated 5 times as the attack process on test nodes is very time-consuming.

\textit{What self-supervision task helps defend which types of graph attacks and why?} In Tables \ref{experiments_adv_cora} and  \ref{experiments_adv_citeseer} we find that introducing self-supervision into adversarial training improves GCN's adversarial defense. (1) Node clustering and graph partitioning are more effective against feature attacks and links attacks, respectively.
During adversarial training, node clustering provides the perturbed feature prior while graph partitioning does perturbed link prior for GCN, contributing to GCN's resistance against feature attacks and link attacks, respectively.  (2)  Strikingly, graph completion boosts the adversarial accuracy by around 4.5 (\%) against link attacks and over 8.0 (\%) against the link \& feature attacks on Cora.  It is also among the best self-supervision tasks for link attacks and link \& feature attacks on Citeseer, albeit with a smaller improvement margin (around 1\%).  In agreement with our earlier conjecture in Section \ref{ss_generalizability}, the topology- and feature-based graph completion constructs (joint) perturbation prior on links and features, which benefits GCN in its resistance against link or link \& feature attacks.

Furthermore, we generate attacks with varying perturbation intensities ($n_\mathrm{perturb} \in \{1,2,3,4\}$) to check the generalizabilty of our conclusions.  Results in Appendix C show that with self-supervision introduced in adversarial training, GCN can still improve its robustness facing various attacks at various intensities.

\subsection{Result Summary}
We briefly summarize the results as follows.  

First, among three schemes to incorporate self-supervision into GCNs, multi-task learning works as the regularizer and consistently benefits GCNs in generalizable standard performances with proper self-supervised tasks.  Pretraining \& finetuning switches the objective function from self-supervision to target supervision loss, which easily ``overwrites'' shallow GCNs and gets limited performance gain.  Self-training is restricted in what pseudo-labels are assigned and what data are used to assign pseudo-labels. And its performance gain is more visible in few-shot learning and can be diminishing with slightly increasing labeling rates.  

Second, through multi-task learning, self-supervised tasks provide informative priors that can benefit GCN in generalizable target performance.  Node clustering and graph partitioning provide priors on node features and graph structures, respectively; whereas graph completion with (joint) priors on both help GCN in context-based feature representation.
Whether a self-supervision task helps a SOTA GCN in the standard target performance depends on whether the dataset allows for quality pseudo-labels corresponding to the task and whether self-supervised priors complement existing architecture-posed priors.
\begin{table}[!t]
\scriptsize
\begin{center}
\caption{Adversarial defense performances on Cora using adversarial training (abbr. AdvT) without or with graph self-supervision.  Attacks include those on links, features (abbr. Feats), and both.  \darkred{Red} numbers indicate the best two performances in each attack scenario (node classification accuracy; unit: \%).}
\label{experiments_adv_cora}
\resizebox{0.48\textwidth}{!}{
\begin{tabular}{c | c | c | c | c}
    \hline\hline
     Attacks & None & Links & Feats & Links \& Feats \\
    \hline
    \hline
    GCN & \darkred{80.61} $\pm$ 0.21 & 28.72 $\pm$ 0.63 & 44.06 $\pm$ 1.23 & 8.18 $\pm$ 0.27 \\ \hline
    AdvT & 80.24 $\pm$ 0.74 & 54.58 $\pm$ 2.57 & 75.25 $\pm$ 1.26 & 39.08 $\pm$ 3.05 \\  
    \hline
    AdvT+Clu & 80.26 $\pm$ 0.99 & 55.54 $\pm$ 3.19 & \darkred{76.24} $\pm$ 0.99 & \darkred{41.84} $\pm$ 3.48 \\
    AdvT+Par & \darkred{80.42} $\pm$ 0.76 & \darkred{56.36} $\pm$ 2.57 & 75.88 $\pm$ 0.72 & 41.57 $\pm$ 3.47 \\
    AdvT+Comp & 79.64 $\pm$ 0.99 & \darkred{59.05} $\pm$ 3.29 & \darkred{76.04} $\pm$ 0.68 & \darkred{47.14} $\pm$ 3.01 \\
    \hline
\end{tabular}}
\vspace{-0.5em}
\end{center}
\end{table}

\begin{table}[!t]
\scriptsize
\begin{center}
\caption{Adversarial defense performances on Citeseer using adversarial training without or with graph self-supervision.}
\label{experiments_adv_citeseer}
\resizebox{0.48\textwidth}{!}{
\begin{tabular}{c | c | c | c | c}
    \hline\hline
     Attacks & None & Links & Feats & Links \& Feats \\
    \hline
    \hline
    GCN & \darkred{71.05} $\pm$ 0.56 & 13.68 $\pm$ 1.09 & 22.08 $\pm$ 0.73 & 3.08 $\pm$ 0.17 \\ \hline 
    AdvT & 69.98 $\pm$ 1.03 & 39.32 $\pm$ 2.39 & 63.12 $\pm$ 0.62 & 26.20 $\pm$ 2.09 \\
    \hline
    AdvT+Clu & \darkred{70.13} $\pm$ 0.81 & 40.32 $\pm$ 1.73 & \darkred{63.67} $\pm$ 0.45 & 27.02 $\pm$ 1.29 \\
    AdvT+Par & 69.96 $\pm$ 0.77 & \darkred{41.05} $\pm$ 1.91 & \darkred{64.06} $\pm$ 0.24 & \darkred{28.70} $\pm$ 1.60 \\
    AdvT+Comp & 69.98 $\pm$ 0.82 & \darkred{40.42} $\pm$ 2.09 & 63.50 $\pm$ 0.31 & \darkred{27.16} $\pm$ 1.69 \\
    \hline
\end{tabular}}
\vspace{-0.5em}
\end{center}
\end{table}

Last, multi-task self-supervision in adversarial training improves GCN's robustness against various graph attacks.  Node clustering and graph partitioning provides priors on features and links, and thus defends better against feature attacks and link attacks, respectively. Graph completion, with (joint) perturbation priors on both features and links, boost the robustness consistently and sometimes drastically for the most damaging feature \& link attacks.  

\section{Conclusion}
\label{conclusion}
In this paper, we present a systematic study on the standard and adversarial performances of incorporating self-supervision into graph convolutional networks (GCNs).
We first elaborate three mechanisms by which self-supervision is incorporated into GCNs and rationalize their impacts on the standard performance from the perspective of optimization.  Then we focus on multi-task learning and design three novel self-supervised learning tasks.  And we rationalize their benefits in generalizable standard performances on various datasets from the perspective or data-driven regularization.  Lastly, we integrate multi-task self-supervision into graph adversarial training and show their improving robustness of GCNs against adversarial attacks.  Our results show that, with properly designed task forms and incorporation mechanisms, self-supervision benefits GCNs in gaining both generalizability and robustness.  Our results also provide rational perspectives toward designing such task forms and incorporation tasks given data characteristics, target tasks and neural network architectures.  

\section*{Acknowledgements}

We thank anonymous reviewers for useful comments that help improve the paper during revision.  This study was in part supported by the National Institute of General Medical
Sciences of the National Institutes of Health [R35GM124952 to Y.S.], and a US Army Research Office Young Investigator Award [W911NF2010240 to Z.W.].

\bibliography{example_paper}
\bibliographystyle{icml2020}

\end{document}